\documentclass[manuscript,screen]{acmart}

\AtBeginDocument{%
  \providecommand\BibTeX{{%
    \normalfont B\kern-0.5em{\scshape i\kern-0.25em b}\kern-0.8em\TeX}}}

\copyrightyear{2026} 
\acmYear{2026}
\setcopyright{none}
\acmConference[FAccT '26]{The 2026 ACM Conference on Fairness, Accountability, and Transparency}{June 27--30, 2026}{Montreal, Canada}
\acmBooktitle{The 2026 ACM Conference on Fairness, Accountability, and Transparency (FAccT '26), June 27--30, 2026, Montreal, Canada}

\usepackage{graphicx}   
\usepackage{amsmath}    
\usepackage{algorithm}

\begin{document}
\providecommand{\DIFadd}[1]{}
\providecommand{\DIFdel}[1]{}
\renewcommand{\DIFadd}[1]{{\color{red}#1}}
\renewcommand{\DIFdel}[1]{}

\title[FairTree: Subgroup Fairness Auditing]{FairTree: Subgroup Fairness Auditing of Machine Learning Models with Bias--Variance Decomposition}
\titlenote{Accepted at the 2026 ACM Conference on Fairness, Accountability, and Transparency (FAccT '26), June 25--28, 2026, Montr\'eal, Canada. This is the author's preprint version.}
\renewcommand{\shorttitle}{FairTree: Subgroup Fairness Auditing}

\author{Rudolf Debelak}
\orcid{https://orcid.org/0000-0001-8900-2106}
\affiliation{%
  \institution{University of Zurich}
  \city{Zurich}
  \country{Switzerland}
}
\affiliation{%
  \institution{EPFL}
  \city{Lausanne}
  \country{Switzerland}
}
\email{rudolf.debelak@epfl.ch}

\begin{abstract}
  The evaluation of machine learning models typically relies mainly on performance metrics based on loss functions, which risk to overlook changes in performance in relevant subgroups. Auditing tools such as SliceFinder and SliceLine were proposed to detect such groups, but usually have conceptual disadvantages, such as the inability to directly address continuous covariates. In this paper, we introduce FairTree, a novel algorithm adapted from psychometric invariance testing. Unlike SliceFinder and related algorithms, FairTree directly handles continuous, categorical, and ordinal features without discretization. It further decomposes performance disparities into systematic bias and variance, allowing a categorization of changes in algorithm performance. We propose and evaluate two variations of the algorithm: a permutation-based approach, which is conceptually closer to SliceFinder, and a fluctuation test. Through simulation studies that include a direct comparison with SliceLine, we demonstrate that both approaches have a satisfactory rate of false-positive results, but that the fluctuation approach has relatively higher power. We further illustrate the method on the UCI Adult Census dataset. The proposed algorithms provide a flexible framework for the statistical evaluation of the performance and aspects of fairness of machine learning models in a wide range of applications even in relatively small data.
\end{abstract}




\maketitle

\section{Introduction}
The evaluation of model performance in machine learning  is a complex, multi-faceted problem. An aspect that is closely related to the evaluation of fairness is the detection of subpopulations where a chosen machine learning model performs worse than in other parts of the population. This problem is particularly relevant in prediction problems, where such algorithms can help with the detection of use cases where models predictions are found to be systematically biased or unreliable as an aspect of model validation.

The aim of this paper is the presentation of a new method named FairTree that is based on recursive partitioning, a method that was adapted from psychometrics and econometrics \cite{merkle2013, merkle2014, zeileis2007}. Compared to existing frameworks, our approach has the following features: a) It can handle ordered and unordered categorical features as well as continuous features without requiring their discretization. This reduces the risk of overlooking slices with poor prediction performance. b) In the interpretation of the slices, we distinguish between effects on bias and variance, since those lead to qualitative differences in their interpretation. While bias detection reveals if and where a machine learning model systematically discriminates against a group (e.g., consistently underpredicting income for women), detecting variance shifts reveals where a model is unreliable or unstable for a specific group. The proposed method leads to an intuitive visualization that summarizes its findings.  

\section{Related Work}
Model validation is a central part of establishing machine learning pipelines. While general metrics such as the root mean squared error can provide guidance on the overall performance of a model, practitioners and researchers are also interested in finding populations where a model shows exceptionally bad performance, which also includes systematic biases. This type of evaluation is closely related to the assessment of a model's fairness and the possible discrimination of relevant subpopulations. In the following two subsections, we will first briefly summarize work in this line of research in machine learning research, before describing related approaches in psychometrics.

\subsection{Fairness and Evaluation in Machine Learning}
The evaluation of fairness in machine learning presents a rich, complex challenge. At its core is the problem of how ethical and legal frameworks for fairness can be formalized \cite{mehrabi2021}. Chouldechova and G'Sell \cite{chouldechova2017} propose a recursive partitioning framework, also based on score-type tests in the tradition of Zeileis et al. \cite{zeileis2008}, for identifying subgroups where two classification models differ in their fairness properties, such as disparities in false positive rates. While their approach shares the same methodological foundations as FairTree, it addresses a different problem: comparing fairness metrics across two models rather than auditing the performance of a single model. FairTree instead targets the loss function of a single model directly, applies to both classification and regression settings, and decomposes detected performance disparities into bias and variance components. Other approaches for evaluating fairness include structured regression for intersectional subgroup analysis \cite{herlihy2024} and complementary interactive visualization frameworks \cite[e.g.,][]{cabrera2019}.

The FairTree framework operates in the domain of group fairness metrics by comparing the loss function of models across different groups. In contrast to classical metrics for assessing fairness in machine learning, such as demographic parity or equalized odds, FairTree is not based on a comparison of two predefined groups. Kearns et al. \cite{kearns2018} address a related problem by auditing fairness over rich collections of subgroups defined by conjunctions of features, but their combinatorial approach assumes discrete features and focuses on learning fair classifiers rather than diagnosing the source of performance disparities. FairTree instead proposes a method of interpretable machine learning to detect groups of data points, which are defined by values of observed covariates, between which the model shows performance heterogeneity.  Methods that are based on the comparison of pre-existing groups can overlook a change in the performance of models by comparing groups which are not related to observed model changes. The proposed algorithm aims to address this drawback by detecting groups which are affected by changes in the predictive performance of a machine learning model, as measured by the loss function in the validation or test data set.

Several approaches for detecting data slices with poor performance exist. A general approach consists of clustering data points that show similar values in their loss function. An important alternative method is the SliceFinder framework \cite{chung2019a, chung2019b}, which aims at finding clusters of data points which not only share a similar level of their loss function, but are also interpretable for humans. By design, this framework searches for groups of data points which a) have a specific minimum effect size, b) are statistically significant, and c) can be described by a minimum number of literals, or defining characteristics (such as "sex", "education", or similar). To find such slices, Chung et al. \cite{chung2019a} propose two automated methods, with the first being based on decision trees and the second one being based on lattice searching. The first approach leads to an exhaustive partitioning of the feature space into slices. The second approach is relatively more expensive and leads to slices that can potentially overlap. Both approaches can be directly applied with categorical features. Numerical features can be handled as well by discretizing them. If an unfortunate discretization is chosen, it might lead to a misinterpretation of effects, and in the worst case, even to overlooking small slices with poor performance.

While the SliceFinder approach is conceptually interesting and related to the presented FairTree approach, there appears to be no current implementation of this approach in software. More recently, the  SliceLine algorithm was proposed as a computationally more efficient alternative \cite{sagadeeva2021}, which is also available via a Python package. However, this approach makes the assumption that the underlying covariates are categorical.

\subsection{Psychometric Measurement Theory and Fairness}
Psychometrics has a strong and long tradition of measurement theory, which also affected research in artificial intelligence \cite{bringsjord2003, salaudeen2025, wang2023, xiao2023, xu2022, zhuang2025}. This field of research is typically concerned with the modeling of responses of test takers to cognitive tasks (or items). The pattern of observed correct and incorrect tasks is used to estimate the overall ability of test takers, which is interpreted as a latent variable.

Another important research question in psychometrics concerns the problem of whether predictive models are stable over relevant subpopulations \cite{millsap1997, millsap2007}. An important special case of this research question is differential item functioning, which is the relative change of item characteristics, like difficulty, across groups. For instance, if an item is disproportionately difficult for women compared to men of equal ability, ignoring this causes biased test results and, as a consequence, discrimination. 

To detect differential item functioning, several techniques have been developed. An important  exploratory method are Rasch and PL trees, which aim at detection parameter changes between subpopulations of interest \cite{strobl2015}. The results of this analysis are typically presented as a decision tree, where different nodes present subpopulations with different profiles of models parameters that are based on different combinations of features. An example based on the \texttt{psychotree} package in R \cite{strobl2015} is presented in Figure~\ref{fig:Raschtree} in Appendix~\ref{app:raschtree}. A distinct feature of this method is that it is able to handle continuous, ordered and unordered categorical features simultaneously without requiring discretization of the continuous features. In the next section, we propose a new method similar to Rasch trees that can be used for the evaluation of machine learning models.

\section{The Proposed Method}

\subsection{Problem Formulation}
The method proposed in this paper concerns the following core scenario: We are considering a machine learning model that provides numerical output $Y_{\text{pred}}$ to predict observed values $Y_{\text{obs}}$ based on numerical or verbal input $X$. The instances providing the input can be labeled using categorical, ordinal or continuous covariates $Z$. This scenario includes the following typical applications:
\begin{itemize}
    \item A random forest classifier that decides whether a person has an income above or below \$ 50,000, based on person characteristics such as sex, education or occupation. Here, the outcome is numerical, and the variables defining each person are categorical (sex, occupation) or ordinal (education). This example was used by Chung et al. \cite{chung2019b} and is based on the UCI Census Data.
    \item An automated essay scoring system that provides numerical scores from 1 to 5 for assessing the overall quality of student texts. Here, 1 could label a text of the lowest quality, while 5 would label a text of the highest quality. If the model shows high bias against, for instance, non-native speakers, it systematically lowers their scores. However, if it shows high variance, it treats non-native speakers inconsistently by grading some accurately and others arbitrarily. Both are fairness violations.
    
    In this application, the instances providing the input, which are the texts, can be categorized based on categorical variables (gender, language of the text, native language of the author), ordinal variables (educational level of the author), or continuous variables (age of the author).  
\end{itemize}

In this scenario, we aim to check the following null hypothesis on the numerical outputs: Is the expected accuracy of the machine learning model, as measured by the distribution of the loss function in the test data set, invariant across all covariates labeling the instances?

If this null hypothesis is violated, the method should split the sample of instances into smaller subsamples a) which are either below a pre-defined, minimum sample size or b) for which the null hypothesis is not violated. In the current implementation, there is no minimal sample size for the splits resulting from the variation of FairTree that is based on the permutation test, while there is a minimum sample size of 10 for the splits resulting from the fluctuation test. 

\subsection{The Splitting Criterion: A Permutation or Fluctuation Test}
In its most simple version, the null hypothesis tested with FairTrees states that the expected value and variance of the loss function of a machine learning model is equal for two pre-defined groups. As a first option, this statistical null hypothesis is formally tested via a permutation test, which is a standard tool in statistics. By repeatedly shuffling the group membership of the data points and re-calculating the loss function, we obtain a reference distribution against which the observed value of the loss function is compared. If the observed value is extreme with regard to the reference distribution, we conclude that the two groups differ significantly with regard to their average loss function. In the implementation used in this study, by default 5000 permutations were used.

In more complex scenarios, there can be multiple groups, or data points that are defined by a continuous covariate (e.g., age). These cases can essentially be simplified to a scenario with two groups:

\begin{enumerate}
    \item In a scenario with $k$ categorical groups (with $k$ > 2), a naive approach would consider all possible binary partitions, which would be computationally intensive for a large number of groups. A more efficient approach is the efficient CART splitting method, which consists of three steps: First, we calculate the mean loss for each category. Second, we sort the categories by their mean loss. Third, we only evaluate partitions that result from splitting the groups along this order. Compared to the naive approach, this method only requires a number of comparisons that increases linearly with $k$. 

    \item For continuous covariates, all unique split points between consecutive unique values are tested. The permutation test relies on the assumption of exchangeability of the loss values under the null hypothesis: if the model performs equally well across all levels of a covariate, then the assignment of observations to covariate levels is uninformative with respect to the loss, and permuting these assignments yields a valid reference distribution. This assumption is satisfied whenever the observations in the test set are independent and the loss function is well-defined for all observations. No distributional assumptions on the loss values are required.
\end{enumerate}

As a second option, we implement a fluctuation test, specifically adapted from the generalized M-fluctuation test framework commonly used in econometrics for parameter stability checking \cite{zeileis2007}. Unlike the permutation test, which requires computationally expensive resampling for every potential split point, the fluctuation test assesses the stability of the model's loss function analytically.

Formally, we distinguish between systematic bias and variance instability by applying the fluctuation test to two distinct sequences derived from the model's predictions. Let $R_i = Y_{\text{pred}, i} - Y_{\text{obs}, i}$ denote the raw residual for the $i$-th instance. To test whether the model exhibits systematic bias with respect to a continuous covariate $Z$, we first order the $n$ observations according to the values of $Z$ such that $z_{(1)} \le z_{(2)} \le \dots \le z_{(n)}$.

We define the empirical fluctuation process for bias, $W_{\text{bias}}(t)$, based on the cumulative sum (CUSUM) of the centered residuals. Let $\bar{R}$ be the global mean residual over all $n$ observations. The process is defined as \cite{merkle2013}:

\begin{equation}
    W_{\text{bias}}(t) = \frac{1}{\hat{\sigma}_R\sqrt{n}} \sum_{i=1}^{\lfloor nt \rfloor} (R_{(i)} - \bar{R}), \quad 0 \le t \le 1
\end{equation}

where $\hat{\sigma}_R$ is a consistent estimator of the standard deviation of the residuals, and the index $i$ corresponds to the ordering induced by $Z$.

Under the null hypothesis of no systematic bias along $Z$, the Functional Central Limit Theorem (FCLT) states that $W_{\text{bias}}(t)$ converges in distribution to a standard Brownian bridge $W^0(t)$ as $n \to \infty$ \cite{zeileis2007}. To detect a significant deviation, we use the maximum statistic ($S_{\text{bias}}$), which corresponds to the maximum absolute value of the fluctuation process:

\begin{equation}
    S_{\text{bias}} = \max_{i=1, \dots, n} \left| W_{\text{bias}} \left( \frac{i}{n} \right) \right|
\end{equation}

To assess changes in variance, we apply a second fluctuation test to the squared centered residuals. To prevent a shift in the mean from artificially inflating the variance estimates, we utilize conditional centering. If the bias test identifies a significant structural break at index $k$, we center the residuals using their respective group means before and after $k$. If no significant bias is detected, we use global centering.

The empirical fluctuation process for variance, $W_{\text{var}}(t)$, is then computed on the squared conditionally centered residuals $V_{(i)} = (R_{(i)}^*)^2$:

\begin{equation}
    W_{\text{var}}(t) = \frac{1}{\hat{\sigma}_V\sqrt{n}} \sum_{i=1}^{\lfloor nt \rfloor} (V_{(i)} - \bar{V}), \quad 0 \le t \le 1
\end{equation}

where $\bar{V}$ is the mean of $V$ and $\hat{\sigma}_V$ is the standard deviation of $V$. The test statistic $S_{\text{var}}$ is derived by taking the maximum absolute value of this process.

Please note that this process describes a simple univariate stochastic process. The significance of this statistic can be computed using the known boundary crossing probabilities of the Brownian bridge \cite{zeileis2007}, which is given by the Kolmogorov distribution. If $S$ exceeds the critical value defined by the significance level $\alpha$ (e.g., $1.358$ for $\alpha=0.05$), we reject the null hypothesis. In principle, the statistical significance can be tested by other means, such as bootstrapping. The fluctuation test requires that the loss values have finite variance and that the observations are independent conditional on their ordering by the covariate. Under these conditions, which are satisfied for standard loss functions applied to i.i.d. test data, the Functional Central Limit Theorem guarantees that $W_{bias}(t)$ converges in distribution to a standard Brownian bridge as $n \to \infty$. The critical values of the test statistic $S$ then follow from the Kolmogorov distribution, providing an analytical alternative to resampling.

So far, we discussed the fluctuation test when testing for invariance with regard to continuous variables. When testing for invariance with regard to categorical variables, the following approximate approach is suggested: First, we consider the mean residuals (average loss function) for each category. Second, we re-order the categories with regard to their mean residuals. Third, the individual observations in the dataset are sorted according to this new order of their category, and a fluctuation test as outlined for the case of continuous variables is carried out. In case of unordered categorical covariates with multiple classes, 50\% of the data is used to calculate mean residuals and establish an ordering of the categories. The remaining 50\% are used for calculating the fluctuation test to prevent data leakage.

Compared to the first approach based on the permutation test, this approach offers two distinct advantages. First, in settings with continuous covariates, it avoids the multiple testing problem inherent in checking every possible cut-point, which could lead to a high rate of false-positive results, as it evaluates the entire process simultaneously. Second, it is computationally efficient (in a magnitude of $O(n \log n)$ due to sorting), making it suitable for larger datasets where permutation testing is prohibitive.

\subsection{Decomposing Performance: Bias and Variance}
A significant result in the FairTree algorithm indicates a different distribution of the loss function in at least two groups that can be defined based on the observed categorical or continuous covariates. This typically indicates at least one of two problems with the predictions. 

First, the predictions might have a higher variance in one group compared to the other, which leads to more accurate predictions in one group. For instance, one might observe that the variance of the differences between true and predicted values is 10 in one group, but 1 in the other group. This corresponds to a difference in the variance.

Second, the predictions might be significantly higher or lower compared to the true values in one group compared to the other. For instance, the average values of the differences between the true and predicted values might be close to 0 in one group, but close to 1 in the other group. This indicates a difference in the bias. For some widely used loss functions like the mean squared error, there are well-known relationships between the loss function, the bias and the variance. 

In fairness terms, shifts in bias and variance are strongly related to violations of fairness. A significant shift in bias between groups typically indicates a violation of calibration. A significant shift in variance, on the other hand, indicates differential reliability. Even if the average prediction is unbiased, high variance in a protected group means the model has failed to learn the features for that group as effectively as for the majority.

For both proposed variations of the FairTree algorithm we implemented rules to detect whether a shift in the loss function is related to a change in bias or variance. 
This allows a more specific evaluation of the difference in behavior of the machine learning model in both groups.

\subsubsection{Detection of Bias and Variance Shifts in the Permutation-Based Algorithm}
If the permutation-based test detects a systematic change in the loss function of a machine learning model, the sample can get split into two subsamples based on the point of the largest change. It is then possible to calculate the differences in bias and variance of the loss function in both subsamples. A heuristic solution to decide whether the difference between both groups is related to the bias or the variance is based on the comparison of their relative differences: If the absolute difference in the bias is 1.5 times or more larger than the absolute difference in the variance between both groups, the difference can be interpreted as being a difference in the bias. If the absolute difference in the variance is 1.5 times or more larger than the absolute difference in the bias between both groups, the difference can be interpreted as being a difference in the variance. In all other cases, the difference is related to differences in the bias and variance. 

\subsubsection{Detection of Bias and Variance Shifts in the Fluctuation Test}
In the fluctuation test, a sequential procedure is applied to decide whether a difference in the loss function is related to a difference in the bias or variance.

If the outlined fluctuation test leads to a statistically significant results ($p < 0.05$), the sample is first split into two subsamples, based on the point where the process outlined in Equation (2) deviates most strongly from 0. In each of the two resulting subgroups, the loss function is centralized by subtracting the group-wise mean. Otherwise, the loss-function is centered by subtracting the global mean, leading to residuals $r$ with a mean of 0. We now consider the following second process besides $W_{bias}(t)$:

\begin{equation}
    W_{var}(t) = \frac{1}{\hat{\tau}\sqrt{n}} \sum_{i=1}^{\lfloor nt \rfloor} (L^2_{(i)} - \bar{L}^2_{(i)}), \quad 0 \le t \le 1
\end{equation}

Again, $\hat{\tau}$ is an estimate of the variance of the stochastic process and aims to standardize it. Analogous the bias testing process $W_{bias}(t)$ and its resulting p-value $p_{bias}$, this process yields a statistical test $W_{var}(t)$, leading to a p-value $p_{var}$. By checking the significance of both statistical p-value $p_{bias}$ and $p_{var}$ against a Bonferroni-corrected alpha level (e.g., 0.05 / 2 = 0.025), it is tested whether there is a significant difference in bias, variance, neither or both.

\subsection{The Recursive Partitioning Algorithm}
\begin{figure}[!ht]
    \centering
    \includegraphics[scale = 0.65]{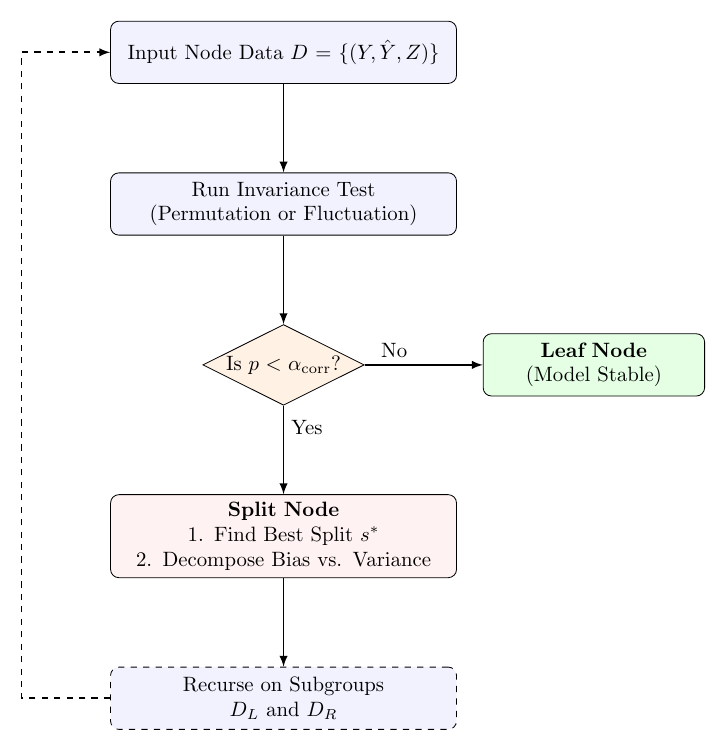}
    \caption{The FairTree recursive partitioning workflow. The algorithm tests for global parameter invariance before splitting and diagnosing the source of error (Bias vs. Variance).}
    \Description{A flow chart indicating the workflow of the recursive partitioning method used in the FairTree algorithm.}
    \label{fig:methodology}
\end{figure}

As an addition, we use a recursive partitioning algorithm to process multiple covariates simultaneously. This aims at detecting covariates that are most strongly related to observed differences in loss functions. The general idea is to first check whether there are significant differences with regard to any of the observed person covariates. If this is the case, we carry out a split at the splitting point that shows the highest difference between both groups, as measured by the p-value of a statistical test. To account for multiple statistical testing, the proposed algorithm carries out a Bonferroni alpha correction before checking if any split has a p-value below a pre-defined level of significance. After the split, which leads to two subsamples, we continue with an evaluation in each subsample until a stopping criterion is reached. This corresponds to standard behavior of comparable algorithms in machine learning, such as decision trees. Two possible stopping criteria are: a) There is no splitting point in the sample that leads to a significant difference between both groups. b) The sample size is below a threshold so that further splitting is not considered meaningful. As was already noted, we used either a permutation test or a fluctuation test to test the null hypothesis of invariance. Figure~\ref{fig:methodology} provides a visualization of the workflow.

\section{An Evaluation with Simulation Studies}
 
To evaluate the described algorithm, we carried out three simulation studies that aimed at evaluating distinct aspects of the method in an abstract scenario. We will describe the setup and the results of these simulation studies in the following subsections. To evaluate the proposed algorithms, we report the following metrics across simulation conditions. \textit{Power} denotes the proportion of simulated data sets in which the algorithm correctly rejects the null hypothesis of invariance when a true effect is present. \textit{Precision} denotes the proportion of detected effects where the algorithm correctly identifies the covariate that is truly associated with the performance disparity. In Simulation Study III, we additionally report the \textit{detection rate}, which is the proportion of simulations in which any covariate is flagged, and the \textit{correct detection rate}, which is the proportion in which all truly affected covariates are correctly identified. For conditions with no simulated effect ($\delta = 0$), rejection rates correspond to the empirical Type I error rate.
 
The scenario depicted in these simulations is inspired by the essay scoring example mentioned in the introduction and aims at the evaluation of the algorithm in comparatively small samples, which would allow an inexpensive detection of changes in bias or variance. We consider a sample of data points, for which we observe pairs of scores that are drawn as integers from an uniform distribution between 1 and 4. For each observed value, we also obtain a prediction, which is generally normal distributed around the observed value, but rounded to the nearest integer. In the simulations, we use the mean squared error over both predictions as loss function and apply the FairTree algorithm which is either based on the fluctuation test or the permutation test.
 
We additionally compared both FairTree variants against SliceLine \citep{sagadeeva2021}, a method for identifying data subgroups with elevated model loss. SliceLine requires a threshold parameter $\alpha_{\text{SL}}$ that controls the minimum relative slice effect size. We evaluated SliceLine with its default threshold ($\alpha_{\text{SL}} = 0.60$) and a tuned configuration ($\alpha_{\text{SL}} = 0.90$) chosen to maximize detection sensitivity, with $k = 10$ top slices, maximum slice complexity $\ell = 2$, and minimum support of 10. Continuous covariates were discretized into quartiles, as SliceLine requires categorical input. SliceLine receives the row-wise mean squared error, which is a composite unsigned metric that responds to both bias and variance effects.
 
\subsection{Simulation Study I: Type I Error and Power}
In the first simulation study, we investigated the Type I error (i.e., the rate of false-positives) and power (i.e., the rate of true-positives) of the algorithms based on the permutation test and the fluctuation test in three separate simulations with two groups.
 
\subsubsection{Simulation Design}
In the first simulation (1.1), we investigated the Type I error and power of the method to detect fairness violations, as indicated by a systematic difference in the loss function between two groups. We simulated two groups of equal size which had the same variance ($\sigma = 0.5$), but different means in their loss function. There were four sample sizes (100, 200, 500 and 1000) and six effect sizes, which corresponded to the mean difference $\delta$ between both groups ($\delta \in \{0, 0.1, 0.15, 0.2, 0.3, 0.4\}$), leading to 4 x 6 = 24 conditions. Note that the conditions with $\delta = 0$ allows the evaluation of the rate of false-positives, whereas the other conditions allow the evaluation of the rate of true-positives. We simulated 1000 data sets under each condition. In each condition, there were five binary covariates that were tested simultaneously, of which only the first one was related to the observed mean differences. The Bonferroni method was used to control the family-wise Type I error rate.
 
 We evaluated the power (i.e., the rate how often a group difference is detected) and the precision (i.e, the rate how often the true covariate is chosen). We expected the power to increase with both the sample size and the simulated effect size, and the precision to be high ($\geq 0.9$) when effects are detected. 
 
In the second simulation (1.2), we investigated the power of the method when the relative size of both groups varies, for an overall sample size of 500. The aim of this simulation study was to evaluate the sensitivity of the proposed algorithm when one group is significantly smaller than the other one, which is a common scenario in contexts that investigate the detection of discrimination against specific groups. The size of the smaller group was 50, 100, 150, 200 or 250. There was a set difference of $\delta = 0.25$ between the loss functions of both groups, with the minority group showing the larger average loss function.  
 
\subsubsection{Results}
The following Table~\ref{tab:power_precision_N_effect} summarizes the results in Simulation 1.1. Please note that the results for effects of 0 correspond to the rate of false-positive results, which should be close to the nominal alpha level of 0.05. For all other effect sizes, we observe the rate of power, which is the expected rate to detect this effect. The precision denotes the rate with which the covariate that is truly related to bias is correctly detected by FairTree.
 
\begin{table}[ht]
\centering
\caption{Simulation Results 1.1: Power and Precision by Sample Size (N) and Effect Size}
\label{tab:power_precision_N_effect}
\resizebox{\textwidth}{!}{%
\begin{tabular}{rrcccccccc}
\toprule
 & & \multicolumn{2}{c}{Permutation Test} & \multicolumn{2}{c}{Fluctuation Test} & \multicolumn{4}{c}{SliceLine} \\
\cmidrule(lr){3-4} \cmidrule(lr){5-6} \cmidrule(lr){7-10}
 & & & & & & \multicolumn{2}{c}{Default ($\alpha_{\text{SL}}=0.6$)} & \multicolumn{2}{c}{Tuned ($\alpha_{\text{SL}}=0.9$)} \\
\cmidrule(lr){7-8} \cmidrule(lr){9-10}
N & Effect Size & Power & Precision & Power & Precision & Power & Precision & Power & Precision \\
\midrule
100  & 0.0  & 0.044 & -     & 0.020 & -     & 0.000 & - & 0.943 & -     \\
     & 0.1  & 0.040 & 0.200 & 0.034 & 0.382 & 0.000 & - & 0.950 & 0.759 \\
     & 0.15 & 0.054 & 0.296 & 0.055 & 0.673 & 0.000 & - & 0.950 & 0.741 \\
     & 0.2  & 0.056 & 0.464 & 0.111 & 0.811 & 0.000 & - & 0.947 & 0.749 \\
     & 0.3  & 0.072 & 0.583 & 0.411 & 0.971 & 0.000 & - & 0.950 & 0.793 \\
     & 0.4  & 0.176 & 0.841 & 0.746 & 0.995 & 0.000 & - & 0.967 & 0.896 \\
\addlinespace
200  & 0.0  & 0.054 & -     & 0.027 & -     & 0.000 & - & 0.760 & -     \\
     & 0.1  & 0.042 & 0.095 & 0.090 & 0.689 & 0.000 & - & 0.758 & 0.487 \\
     & 0.15 & 0.044 & 0.136 & 0.226 & 0.867 & 0.000 & - & 0.764 & 0.480 \\
     & 0.2  & 0.062 & 0.258 & 0.428 & 0.953 & 0.000 & - & 0.780 & 0.542 \\
     & 0.3  & 0.150 & 0.600 & 0.894 & 0.990 & 0.000 & - & 0.847 & 0.759 \\
     & 0.4  & 0.394 & 0.929 & 0.994 & 1.000 & 0.000 & - & 0.944 & 0.945 \\
\addlinespace
500  & 0.0  & 0.040 & -     & 0.031 & -     & 0.000 & - & 0.256 & -     \\
     & 0.1  & 0.068 & 0.235 & 0.241 & 0.892 & 0.000 & - & 0.251 & 0.279 \\
     & 0.15 & 0.050 & 0.240 & 0.650 & 0.983 & 0.000 & - & 0.282 & 0.372 \\
     & 0.2  & 0.090 & 0.556 & 0.935 & 0.997 & 0.000 & - & 0.339 & 0.537 \\
     & 0.3  & 0.350 & 0.891 & 1.000 & 1.000 & 0.000 & - & 0.656 & 0.921 \\
     & 0.4  & 0.878 & 0.984 & 1.000 & 1.000 & 0.000 & - & 0.958 & 0.991 \\
\addlinespace
1000 & 0.0  & 0.032 & -     & 0.046 & -     & 0.000 & - & 0.032 & -     \\
     & 0.1  & 0.048 & 0.458 & 0.597 & 0.975 & 0.000 & - & 0.035 & 0.457 \\
     & 0.15 & 0.078 & 0.513 & 0.975 & 0.999 & 0.000 & - & 0.062 & 0.565 \\
     & 0.2  & 0.174 & 0.828 & 1.000 & 1.000 & 0.000 & - & 0.138 & 0.841 \\
     & 0.3  & 0.730 & 0.978 & 1.000 & 1.000 & 0.000 & - & 0.656 & 0.995 \\
     & 0.4  & 0.990 & 0.998 & 1.000 & 1.000 & 0.000 & - & 0.987 & 1.000 \\
\bottomrule
\multicolumn{10}{l}{\footnotesize \textit{Note:} SliceLine Default = $\alpha_{\text{SL}} = 0.60$; SliceLine Tuned = $\alpha_{\text{SL}} = 0.90$.} \\
\multicolumn{10}{l}{\footnotesize Power indicates the rate of flagging any covariate; Precision indicates the rate of flagging the true underlying cause.} \\
\end{tabular}%
}
\end{table}
 
In summary, both FairTree algorithms maintain Type I error rates close to the nominal level of 0.05. The fluctuation test consistently shows substantially higher power than the permutation test across all sample sizes and effect sizes, with the difference being most pronounced at moderate effect sizes ($\delta$ = 0.15 to 0.2) where the permutation test has near-zero power while the fluctuation test already achieves meaningful detection rates. The precision of both tests is high when effects are detected, with the fluctuation test reaching precision above 0.9 at smaller effect sizes.
 
SliceLine with the default threshold ($\alpha_{\text{SL}} = 0.60$) did not detect any effect in any of the 24 conditions. With the tuned setting ($\alpha_{\text{SL}} = 0.90$), SliceLine detects effects but with severely inflated false-positive rates. The results show that the optimal $\alpha_{\text{SL}}$ depends on both the sample size and the unknown effect size, making it impractical to tune without prior knowledge of the effect structure.
 
For Simulation Study 1.2, the results in Table~\ref{tab:power_analysis_comparison} were obtained. The fluctuation test maintains high power even when the minority group constitutes only 20\% of the sample, whereas the permutation test shows uniformly low power across all minority proportions.
 
\begin{table}[ht]
    \centering
    \caption{Power Analysis Comparison: Impact of Minority Group Proportion and Size on Test Performance}
    \label{tab:power_analysis_comparison}
    \begin{tabular}{rrrrrr}
        \toprule
        & & \multicolumn{2}{c}{FairTree} & \multicolumn{2}{c}{SliceLine} \\
        \cmidrule(lr){3-4} \cmidrule(lr){5-6}
        Minority Prop. & Minority N & Perm. Test & Fluct. Test & Default & Tuned \\
        \midrule
        0.10 & 50  & 0.192 & 0.315 & 0.000 & 0.052 \\
        0.20 & 100 & 0.282 & 0.908 & 0.000 & 0.031 \\
        0.30 & 150 & 0.370 & 0.991 & 0.000 & 0.089 \\
        0.40 & 200 & 0.384 & 0.998 & 0.000 & 0.213 \\
        0.50 & 250 & 0.342 & 0.999 & 0.000 & 0.362 \\
        \bottomrule
        \multicolumn{6}{l}{\footnotesize \textit{Note:} SliceLine Default = $\alpha_{\text{SL}} = 0.60$; \footnotesize SliceLine Tuned = $\alpha_{\text{SL}} = 0.90$.} \\
    \end{tabular}
\end{table}
 
SliceLine with the default threshold again shows zero detection. Even with the tuned threshold, SliceLine achieves at most 36.2\% detection at balanced groups.
 
In summary, this first evaluation shows that the proposed method has desirable statistical properties under the relatively simple conditions considered here. Under all conditions, the fluctuation test was found to have a higher power than the permutation test, and SliceLine's performance depends on its threshold parameter.
 
\subsection{Simulation Study II: Bias-Variance Classification Accuracy}
A core feature of FairTree is that the algorithm aims to distinguish between changes in bias and variance of the loss function of machine learning systems. In the second set of simulation studies, we evaluated whether this is the case, again in relatively small samples. 
 
\subsubsection{Simulation Design}
This set of simulations aimed at evaluating the qualitative and quantitative assessment of bias and variance that results from the proposed algorithm. In Simulation Study II, we generated data under four scenarios, where two groups differed with regard to their means, variances, both or neither. The sample size was 500, and we considered two groups of equal size. If there were no differences, the error terms were drawn from a distribution with $\mu = 0$ and $\sigma = 0.5$. In case of bias differences, the mean error $\mu$ for the affected group was set to $-\delta$. In case of variance differences, the standard deviation $\sigma$ for the affected group was set to $0.5 + \delta$. We simulated three different effect sizes of $\delta = 0.15$, $\delta = 0.25$ and $\delta = 0.4$, leading to 3 x 4 = 12 different conditions. We simulated 500 data sets under each condition.
 
 
\subsubsection{Results}
The aggregated results of this evaluation for the permutation-based variation of the fluctuation algorithm are presented in the following Table~\ref{tab:confusion_matrices}.
 
\begin{table}[ht]
\centering
\caption{Confusion Matrices of Detected vs. True Issues and Overall Detection Rates}
\label{tab:confusion_matrices}
\resizebox{\textwidth}{!}{%
\begin{tabular}{lrrrrrrrrrr}
\toprule
& \multicolumn{4}{c}{Permutation Test (Classified)} & \multicolumn{4}{c}{Fluctuation Test (Classified)} & \multicolumn{2}{c}{SliceLine (Detection Only)} \\
\cmidrule(lr){2-5} \cmidrule(lr){6-9} \cmidrule(lr){10-11}
True Issue & Bias & Both & Neither & Var. & Bias & Both & Neither & Var. & Default & Tuned \\
\midrule
Bias     & 0.47 & 0.02 & 0.51 & 0.00 & 0.89 & 0.05 & 0.06 & 0.00 & 0.00 & 0.49 \\
Both     & 0.25 & 0.27 & 0.10 & 0.38 & 0.10 & 0.80 & 0.03 & 0.07 & 0.00 & 1.00 \\
Neither  & 0.00 & 0.01 & 0.94 & 0.05 & 0.02 & 0.00 & 0.96 & 0.02 & 0.00 & 0.05 \\
Variance & 0.00 & 0.01 & 0.06 & 0.93 & 0.00 & 0.02 & 0.14 & 0.83 & 0.00 & 0.99 \\
\bottomrule
\multicolumn{11}{l}{\footnotesize \textit{Note:} Var. = Variance. Values are averaged across $\delta \in \{0.15, 0.25, 0.4\}$.} \\
\multicolumn{11}{l}{\footnotesize For FairTree variations, rows sum to 1.0. SliceLine cannot classify the type of issue; its columns reflect the raw rate of flagging any covariate.} \\
\end{tabular}%
}
\end{table}
 
As can be seen, data sets where only a difference in bias, in variance or neither was simulated were usually classified correctly. Data sets with both a difference in bias and variance were more often classified incorrectly by the permutation-based algorithm, but classified correctly by the fluctuation algorithm. This difference between both test types can be related to how the two variations of FairTree distinguish between bias and variance. While the variation based on the permutation test is based on a heuristical rule, the variation based on the fluctuation test is based on statistical p-values, and therefore sensitive to the overall sample size underlying the statistical tests. 
 
Because SliceLine does not distinguish between bias and variance, it can only be evaluated on detection in this study. SliceLine with the default threshold did not detect bias-only effects at any $\delta$, but showed some tendency to detect changes in bias and variance in its tuned version. However, only FairTree provides classification of the detected issue type---a capability that SliceLine's architecture does not support.
 
\subsection{Simulation Study III: Realistic Scenarios}
Simulation Study III evaluated the FairTree algorithm and SliceLine under different realistic conditions. We simulated 200 data sets under each of the various simulated conditions. We considered the following scenarios:
\begin{itemize}
    \item Mixed covariate types: There were six covariates, of which one was binary, two were categorical with three levels, two were categorical with five levels, and one was continuous. Only the binary covariate was related to differences in the loss function, with a mean difference of $\delta = 0.25$. The sample size was 500, the error variance was $\sigma = 0.5$.
    \item Two independent causes: There are four covariates: one binary (gender), one categorical with three levels, one categorical with five levels, and one continuous. Two covariates are related to fairness violations of different types: "Gender" (2 categories) is associated with systematic bias $\delta = 0.2$), while "Topic" is associated with increased variance for a subgroup ($\sigma = 0.8$ compared to $\sigma = 0.5$). The sample size is 800.
    \item Continuous covariate: There are three covariates, and one of them, which is continuous, is related to a difference in bias. The continuous variable can be interpreted as an age variable, which is distributed uniformly between the values of 13 and 18. Below the value of 15, the mean loss function is reduced by $\delta = 0.3$. The simulated sample size is 600, and the error variance is $\sigma = 0.5$.
    \item Categorical with multiple classes: There are three covariates: one categorical with eight levels ("Topic"), one binary ("Gender"), and one categorical with three levels. Only two categories of "Topic" are associated with bias, showing a mean difference of $\delta = 0.3$. This scenario tests the approach for finding the optimal partition of multi-class categoricals. The sample size is 700, and the error variance is $\sigma = 0.5$.
    \item Interaction effect: There are three covariates: two binary ("Gender" and "Topic") and one categorical with three levels. There is an interaction effect: only boys writing on one topic are penalized ($\delta = 0.35$), while boys writing on the other topic and girls on either topic show no bias. This scenario tests whether FairTree can detect violations that manifest only in specific subgroups. The sample size is 800, and the error variance is $\sigma = 0.5$.
    \item Confounding:  There are three categorical covariates with two, three and three categories, respectively. The covariates "Gender" (2 categories) and "Topic" (3 categories) are correlated. While boys tend to choose Topic A (70\%), girls tend to choose Topic B (70\%). While the bias concerns only the "Topic" covariate (a specific topic is penalized by $\delta = 0.3$), it is assessed whether the FairTree algorithm also incorrectly detects "Gender" as biased due to the confounding. The sample size is 600.
    \item Small Minority: There are three covariates: two binary ("Native Speaker Status" and "Gender") and one categorical with three levels. Non-native speakers constitute a small minority (10\% of the sample) but experience substantial bias ($\delta = 0.5$). This scenario tests FairTree's ability to detect violations affecting small subgroups, where statistical power is reduced due to limited group size. The sample size is 500, and the error variance is $\sigma = 0.5$.
\end{itemize}
 
\subsubsection{Results}
The results for the permutation-based and the fluctuation-based algorithms, as well as for SliceLine, are presented in Table~\ref{tab:simulation_results}. The detection rate provides the rate of the positive results where any covariate is flagged, the correct detection rate provides the rate of correct results where covariates causing bias are correctly detected, and precision is the ratio of true positives among all positives.
 
\begin{table}[ht]
\centering
\caption{Simulation Results: Comparison of FairTree and SliceLine Performance by Scenario}
\label{tab:simulation_results}
\resizebox{0.8\textwidth}{!}{%
\begin{tabular}{lrrrrrrrrr}
\toprule
& \multicolumn{3}{c}{Permutation Test} & \multicolumn{3}{c}{Fluctuation Test} & \multicolumn{3}{c}{SliceLine (Tuned)} \\
\cmidrule(lr){2-4} \cmidrule(lr){5-7} \cmidrule(lr){8-10}
Scenario & Det. & Corr. & Prec. & Det. & Corr. & Prec. & Det. & Corr. & Prec. \\
\midrule
A Single Cause   & 0.32 & 0.18 & 0.59 & 1.00 & 0.99 & 0.99 & 0.41 & 0.40 & 0.96 \\
B Two Causes     & 1.00 & 0.99 & 0.99 & 0.97 & 0.97 & 1.00 & 0.72 & 0.72 & 1.00 \\
C Continuous     & 0.10 & 0.04 & 0.37 & 1.00 & 1.00 & 1.00 & 0.07 & 0.02 & 0.31 \\
D Multiclass     & 0.70 & 0.66 & 0.94 & 0.95 & 0.94 & 0.99 & 0.04 & 0.00 & 0.00 \\
E Interaction    & 0.55 & 0.52 & 0.94 & 1.00 & 1.00 & 1.00 & 0.49 & 0.49 & 1.00 \\
F Confounding    & 0.55 & 0.48 & 0.87 & 1.00 & 0.66 & 0.66 & 0.44 & 0.38 & 0.85 \\
G Small Minority & 0.90 & 0.86 & 0.97 & 0.94 & 0.93 & 0.98 & 0.11 & 0.05 & 0.43 \\
\bottomrule
\multicolumn{10}{l}{\footnotesize \textit{Note:} Det.\ = Detection Rate; Corr.\ = Correct Detection Rate; Prec.\ = Precision.} \\
\multicolumn{10}{l}{\footnotesize SliceLine with $\alpha_{\text{SL}} = 0.90$ is shown. SliceLine Default ($\alpha_{\text{SL}} = 0.60$) detected no effects in any scenario.} \\
\end{tabular}%
}
\end{table}
 
These results indicate that the fluctuation test shows desirable properties in most scenarios, whereas the algorithm based on the permutation test shows comparatively lower power in the analysis of continuous covariates. In the confounding scenario (F), both FairTree variants show reduced precision, reflecting the inherent difficulty of distinguishing a true cause from a correlated confounder. While SliceLine did not detect any effects in its base algorithm, the tuned version of SliceLine typically showed a detection rate and precision that are comparable to those of the permutation test, but below the fluctuation test.

\subsection{Implementation Details}
The code for FairTree was written in Python. The code for running the simulation studies is available in several open Jupyter notebooks, which can be found here: \url{https://osf.io/4ktf7/overview?view_only=5ee89ced26344b2f92c9fcbe205ce021}.
To facilitate adoption of the proposed new method, we plan to release FairTree as an open-source Python package upon publication.

\section{An Empirical Illustration: The UCI Adult Census Income Dataset}
As an additional illustration, we apply both FairTree algorithms to the UCI Adult Census Income Dataset \cite{uci_adult} (n $\approx$ 48,000), which contains 14 demographic and socioeconomic features including age, education level (as a numeric scale), hours worked per week, occupation, marital status, race, sex, and native country. The binary target variable indicates whether an individual's annual income exceeds \$ 50,000. A Random Forest classifier (50 trees, maximum depth 10) was trained on a random 70\% split using all 14 features as predictors. FairTree was then applied to the held-out test set (n $\approx$ 9,000) using four covariates: age and hours-per-week as covariates with artificially injected differential performance (a systematic bias offset for age < 25, and added noise for hours-per-week > 55), education-num as a covariate exhibiting natural differential model behavior, and a randomly permuted control feature as a true-null baseline. This design allows simultaneous evaluation of FairTree's sensitivity to injected bias and variance, its ability to discover naturally occurring disparities, and its Type I error control.

We applied an analysis using the FairTree algorithm based on both the permutation and the fluctuation test, as well as SliceLine. The following steps were carried out: 

We first defined training and test sets. The training set entailed 70 \% of the overall data, the test set about 30 \%. The random forest classifier was trained on the training set and applied to the test set, reaching a prediction accuracy of 85.5 \%. 

As a second step, we changed the predictions of the random forest classifier in the test set to obtain a signal for bias and variance: To simulate bias, a constant error of 0.35 was added to everyone under 25 in the feature age; to simulate variance, random noise was added to everyone working more than 55 hours. The aim of this step was to evaluate the sensitivity of the test against these known effects, which mirrors the procedure of \citet{chung2019b} for evaluating the SliceFinder algorithm.

As a third step, we included a permuted age variable to evaluate the tendency of FairTree to detect false-positive effects.

\begin{figure}[!ht]
    \centering
    \includegraphics[width=1\textwidth]{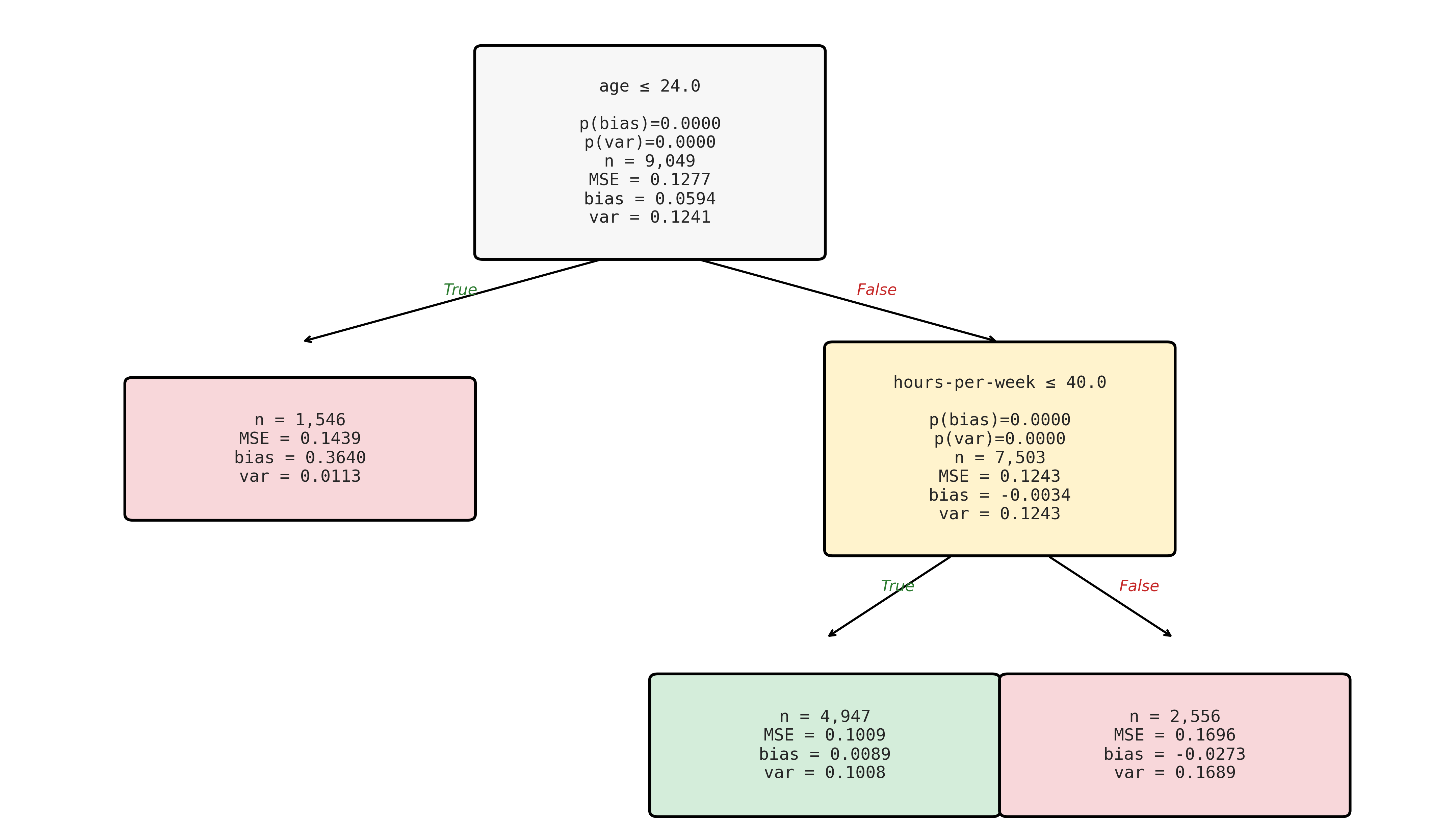}
    \Description{A visualization of the results of the FairTree algorithm in the empirical example via a decision tree.}
    \caption{A visualization of the results of the FairTree algorithm in the empirical example. The sample is first split by age, then by hours-per-week. The green node shows a lower MSE than the two red nodes, which differ with regard to bias and variance.}
    \label{fig:Emp_FairTree}
\end{figure}

For this demonstration, we implement a simplified visualization based on a global priority tree. Rather than recursively determining splits on child nodes, covariates are ranked by their global p-values, mimicking a fairness audit workflow. The results (Table \ref{tab:empirical_results}) show both FairTree algorithms detected changes related to age, hours-per-week, and education. While the permutation test classified the simulated violations correctly, the fluctuation test did not, likely because the large sample size makes even minor baseline deviations statistically significant. However, the fluctuation test was vastly more computationally efficient (21.6 ms vs. 26.4 s). The subpopulation with different model performance are shown in Figure~\ref{fig:Emp_FairTree}. 

Finally, we highlight the negative impact of feature discretization. When \emph{age} and \emph{hours-per-week} were converted into decade bins, the statistical signal was severely diluted because the bins mixed affected and unaffected subjects. This caused FairTree to incorrectly identify the structural break at Age $\le$ 35 instead of the true threshold. Furthermore, when SliceLine with $\alpha_{\text{SL}} = 0.90$ was applied to this discretized data, it missed the injected faults entirely; instead, it extracted an unrelated demographic slice (\emph{sex=Male}). This reflects an natural model error rather than the injected faults. Overall, this underscores the importance of evaluating fairness directly on continuous data. 

\begin{table}[ht]
\centering
\caption{Comparison of the Permutation and Fluctuation Test Performance in the Empirical Illustration}
\label{tab:empirical_results}
\begin{tabular}{lccccc}
\toprule
 & & \multicolumn{2}{c}{\textbf{Fluctuation}} & \multicolumn{2}{c}{\textbf{Permutation}} \\
\cmidrule(lr){3-4} \cmidrule(lr){5-6}
\textbf{Covariate} & \textbf{Expected} & \textbf{Detected} & \textbf{$p$-value} & \textbf{Detected} & \textbf{$p$-value} \\ 
\midrule
Age & Bias & Both & $< 0.001$  & Bias & $0.003$  \\
Hours-per-week & Variance & Both & $< 0.001$  & Variance & $0.001$  \\
Random Control & None & None & $0.505$  & None & $0.768$  \\
Education-Num & Discovery & Both & $< 0.001$  & Both & $0.001$ \\
\bottomrule
\end{tabular}
\end{table}

\section{Discussion}
While machine learning models are usually evaluated by their overall model performance metrics, this practice can overlook weaknesses of the models in specific subgroups. Methods like SliceFinder improve on this practice by comparing the relative model performance in various subgroups. FairTree improves on this approach by integrating a statistically more rigorous approach for handling continuous covariates. In contrast to SliceFinder, the new method also distinguishes between changes in bias and variance, which indicates different types of model inaccuracy and which is supported by intuitive graphical output that is based on decision trees.

Distinguishing these two sources of error is crucial for model checking and remediation. A fairness issue driven by bias might require adapting the loss function or adversarial training. A fairness issue driven by variance, however, often signals a lack of sufficient or high-quality data for that subpopulation, requiring active data collection rather than just algorithmic tuning.

As was shown via simulation studies and empirical applications, the new method has a satisfactory rate of false-positive results and has a high probability to detect violations that are present in the data. This applied to both tests that we used in our simulations, which are a permutation test and a fluctuation test.

The fluctuation test is theoretically closely related to industry checks. In practical application, it might be attractive for real-time fairness monitoring of machine learning systems because of its computational speed. Based on the results of the simulation studies, we recommend the algorithm based on the fluctuation test over that based on the permutation test. The permutation test might be interesting for use cases which include small sample sizes. Here, the permutation test might be used as a follow-up test to confirm the results of the analytical fluctuation test in cases where its analytical approximation via the Functional Central Limit Theorem might be wrong.

\section{Limitations}
The described methods can be extended in various directions, which represent opportunities for further studies: The current method relies on the maximum statistic $D_{max}$, whereas future studies could evaluate methods based on different test statistics. A second extension would be the integration of pruning in the visualization, which could consider whether the found differences in bias and variance are practically significant. Further, the proposed method could be evaluated for a number of additional scenarios that could be inspired by real-world applications of machine learning. 

\section{Ethical Implications}
While algorithms such as FairTree offer a quantitative approach to detect violations of aspects of fairness, it is crucial to reflect on the potential of such methods to obscure inequities. There are several, competing definitions of fairness in machine learning. FairTree is designed to detect fairness violations at the group level; consequently, it does not preclude violations of individual fairness, which require assessment via complementary criteria. Furthermore, there is the potential of selective reporting of results of FairTree (e.g., by choosing a specific loss function) to obscure fairness violations. We advocate to use FairTree as an exploratory tool to detect problems of fairness.

\section{Generative AI usage statement}
Two LLMs (Anthropic Claude Opus 4.5 and 4.6, Google Gemini Pro 3.0 and 3.1) were used to enhance the writing of the texts which were written by the author, to support the coding and to help with the formatting of the tables.


\bibliographystyle{ACM-Reference-Format}
\bibliography{refs.bib} 

\clearpage
\appendix

\section{Appendix A: Illustrative Example of a Rasch Tree}
\label{app:raschtree}

\begin{figure}[!ht]
\centering
\includegraphics[width=1\textwidth]{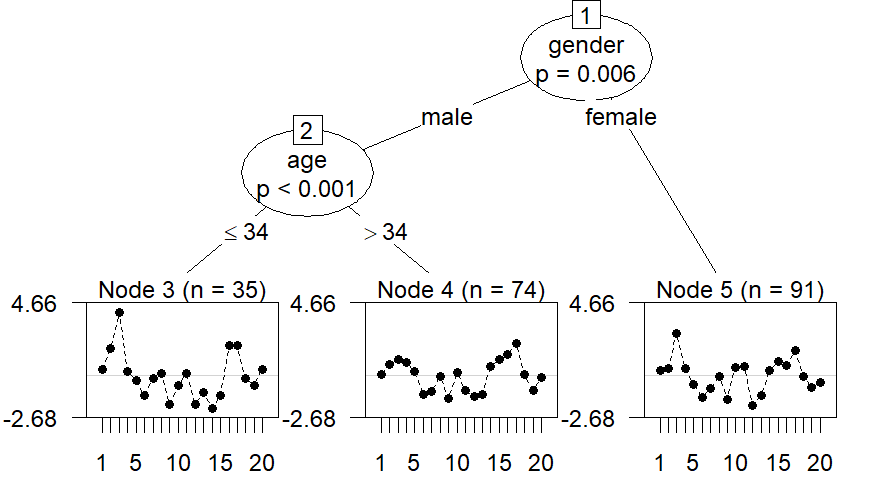}
\Description{A decision tree diagram where the root node splits into multiple branches based on  demographic variables. The nodes show the resulting parameter differences, indicating a model change for different subgroups.}
\caption{An illustrative example of a Rasch Tree, which is used for detecting fairness violations in psychometrics. The illustration shows a shift of model parameters, which are shown by the line plots in the nodes, between relevant subpopulations defined by age and gender.}
\label{fig:Raschtree}
\end{figure}

As discussed in Section 1, Rasch trees are used to detect parameter changes between subpopulations. Figure~\ref{fig:Raschtree} provides a visual example of this method. It shows a change of psychometric item characteristics (i.e., the relative difficulty of several task), which are indicated by line plots in the tree nodes, across three subgroups, which are defined by gender and age (female respondents, male respondents below the age of 34, male respondents above the age of 34).

\end{document}